\documentclass[lettersize,journal]{IEEEtran}
\usepackage[table]{xcolor}
\usepackage{amsmath,amsfonts}
\usepackage{amssymb}
\usepackage{algorithmic}
\usepackage{algorithm}
\usepackage{array}
\ifCLASSOPTIONcompsoc
\usepackage[caption=false, font=normalsize, labelfont=sf, textfont=sf]{subfig}
\else
\usepackage[caption=false, font=footnotesize]{subfig}
\fi
\usepackage{textcomp}
\usepackage{stfloats}
\usepackage{url}
\usepackage{verbatim}
\usepackage{graphicx}
\usepackage{bbding}
\usepackage{booktabs}
\usepackage{multirow}
\usepackage{cite}
\usepackage[%
    colorlinks=true,
    pdfborder={0 0 0},
    linkcolor=red
]{hyperref}
\makeatletter
\newcommand*{\rom}[1]{\expandafter\@slowromancap\romannumeral #1@}
\newcommand{\etal}{\textit{et al}.}
\makeatother
\begin{document}

\title{TMHOI: Translational Model for Human-Object Interaction Detection}

\author{\IEEEauthorblockN{Lijing Zhu\textsuperscript{1}, Qizhen Lan\textsuperscript{1}, Alvaro Velasquez\textsuperscript{2}, Houbing Song\textsuperscript{3}, Acharya Kamal\textsuperscript{4}, Qing Tian\textsuperscript{1} Shuteng Niu\textsuperscript{1}\\}
\IEEEauthorblockA{\textit{\textsuperscript{1} Department of Computer Science, Bowling Green State University}\\
\textit{\textsuperscript{2} Department of Computer Science, University of Colorado Boulder}\\
\textit{\textsuperscript{3} Department of Information Systems, University of Maryland, Baltimore County}\\
\textit{\textsuperscript{4} Department of Electrical Engineering and Computer Science, Embry-Riddle Aeronautical University}}
}


\maketitle

\begin{abstract}

Detecting human-object interactions (HOIs) is a challenging problem in computer vision. Existing techniques for HOI detection heavily rely on appearance-based features, which may not capture other essential characteristics for accurate detection. Furthermore, the use of transformer-based models for sentiment representation of human-object pairs can be computationally expensive. To address these challenges, we propose a novel graph-based approach, TMHOI (Translational Moddel for Human-Object Interaction Detection), that effectively captures the sentiment representation of HOIs by integrating both spatial and semantic knowledge. In a graph, TMHOI takes the components of interaction as nodes, and the spatial relationships between them as edges. Our approach employs a spatial encoder and a semantic encoder to extract spatial and semantic information, respectively, and then combines these encodings to create a knowledge graph that captures the sentiment representation of HOIs. Compared to existing techniques, TMHOI is computationally efficient and allows for the incorporation of prior knowledge, making it practical for use in real-world applications. We demonstrate the effectiveness of our proposed method on the widely-used HICO-DET datasets, where it outperforms existing state-of-the-art graph-based methods by a significant margin. Our results indicate that the TMHOI approach has the potential to significantly improve the accuracy and efficiency of HOI detection, and we anticipate that it will be of great interest to researchers and practitioners working on this challenging task.
\end{abstract}


\begin{figure}[!t]
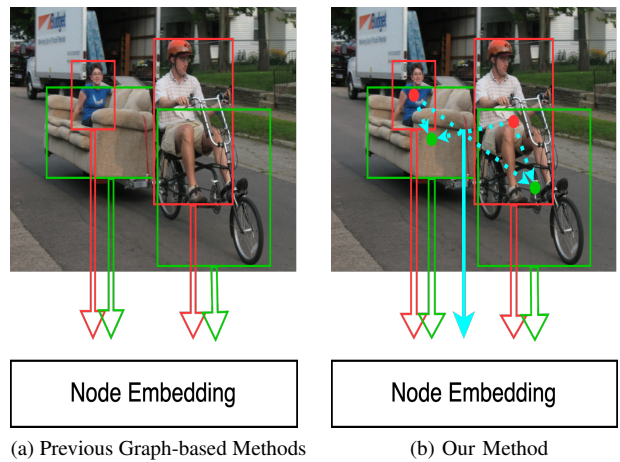

\centering
\subfloat[Previous Graph-based Methods]{\includegraphics[width=1.5in,height = 2.2in]{comparsion_a.pdf}
\label{fig_first_case}}
\hfil
\subfloat[Our Method]{\includegraphics[width=1.5in,height = 2.2in]{comparsion_b.pdf}
\label{fig_second_case}}
\caption{This is the comparison of the node embedding between previous graph-based approaches and our approach. The detected instances are represented by a colorful rectangular bounding box, with a central dot marking the center of the box. A dotted line connects the center point of the bounding box to indicate the relationship between a pair of human and object in the scene. Figure (a) solely relies on the detected bounding box's appearance feature, whereas figure (b) incorporates both the appearance and the relation features of the human and object pair.}
\label{fig_sim}
\end{figure}

\section{Introduction}
In recent years, the development of human-object interaction (HOI) detection has been a rapidly growing field in computer vision. The HOI detection technology is used to recognize the interactions between humans and objects in a given image, allowing for the more comprehensive understanding of human behavior and its relation to the surrounding environment. The development of HOI detection has been driven by advances in deep learning and neural networks, which have enabled more accurate and efficient analysis of visual data. The goal of HOI detection is to identify human-object interactions, which are depicted as the anticipated set $\langle$\textit{person, verb, object}$\rangle$ triplet, for instance, $\langle$\textit{person, read, book}$\rangle$. The detection of HOIs is commonly tackled as a multi-label classification task, whereby an image may entail multiple interactions taking place concurrently. For instance, a single image may exhibit a person engaging in both phone usage and coffee consumption simultaneously, thereby requiring the HOI detector to accurately identify and associate these interactions with the appropriate objects depicted in the image. Detecting the objects and humans present in an image is not the only aspect of the task; recognizing the particular interactions transpiring between them makes it a challenging and intricate undertaking.

As it stands, multiple models and frameworks for human-object interaction (HOI) detection have been proposed. These methods can be broadly classified into one-stage and two-stage approaches. In the case of one-stage approaches\cite{chen2021reformulating},\cite{wang2020learning},\cite{zou2021end},\cite{liao2022gen},\cite{ma2023fgahoi}, the task is accomplished in a single step without the need for detecting the instance first. They can simultaneously detect all pairs of instances on an image and the interaction between these instance pairs. One-stage HOI methods are primarily of two types: interaction-point techniques \cite{liao2020ppdm},\cite{zhong2021glance} and transformer-based methods\cite{tamura2021qpic},\cite{chen2021qahoi}. Interaction-point techniques leverage interaction points to illustrate the region of interaction between humans and objects. This approach efficiently explains the intimate association between people and objects. However, its effectiveness diminishes as the distance between them increases. As for the transformer-based method, it attracts enough attention since the DETR\cite{carion2020end} developed. The DETR introduced the structure resembling transformers into the computer vision algorithms. Transformer-based models have a more intricate architecture and utilize attention mechanisms to facilitate the learning of the head, although at a higher computational expense due to global feature processing. This methodology accommodates complex interactions and produces superior accuracy compared to interaction-point methods, albeit at a slower convergence rate.  

As for the two-stage approaches\cite{qi2018learning},\cite{ulutan2020vsgnet},\cite{zhang2021spatially},\cite{zhou2019relation},\cite{wu2022mining}, the detector is responsible for identifying all feasible instances present in the image. Subsequently, another network performs the HOI detection. The output of the object detector is a set of bounding boxes that localize the detected instances, along with corresponding scores that indicate the confidence score of the instance label. The object detector filters out the instances whose scores are below a certain threshold, retaining only the high-confidence instances. The detected instances from the first stage are used as input to a separate network that performs verb classification tasks. The HOI recognition network typically takes as input appearance features, the visual features of the detected instances, and contextual information such as the spatial layout between human and object pairs. In the previous two-stage approach\cite{chao2018learning},\cite{hou2020visual}, they adopted the multi-stream method, adding a multi-label classifier after the object detector to make action prediction. While the subsequent paper\cite{fang2018weakly},\cite{xiu2018pose} proposes an alternative to the multi-label classifier, such as leveraging pose estimation techniques to incorporate pose information and fuse it with other features to forecast the action category. HOI detection methods often face issues with utilizing a CNN-based backbone feature, which neglects global features. Nonetheless, graph-based methodologies such as those proposed by Li \etal~\cite{li2019transferable}, Ulutan \etal~\cite{ulutan2020vsgnet} and Zhang \etal~\cite{zhang2021spatially} overcome this limitation by capitalizing on appearance features and effectively utilizing spatial features.

A significant proportion of HOI detection methodologies, whether one-stage or two-stage, depend exclusively on appearance features to predict human-object interactions. Nevertheless, such interactions are inherently intricate and markedly variable, making it difficult for HOI detectors to model them with only instance features of human and object entities. To address this challenge, researchers have developed graph-based techniques, which entail constructing a graph model that captures all human-object pair characteristics. By modeling the relationships between all human-object pairs in the image, the graph-based approach can capture spatial and semantic features between pairs of nodes, leading to more accurate HOI detection. In particular, the graph-based approach is useful for modeling complex interactions between multiple humans and objects, which can be difficult to capture with traditional two-stage methods that treat each human-object pair in isolation.   

The advanced graph-based HOI model adopted the appearance feature as the node representation and the spatial feature as the edge representation, then perform the link prediction, which is the verb prediction between the human and the object. Despite the promising performance, this approach only partially explores the semantic information between nodes. Specifically, some existing graph-based methods in HOI detection only leverage instance information and ignore other semantic features hidden in relations between nodes. However, there are certain patterns and tendencies in human-object interactions, such as the fact that people are more likely to read books rather than tables. The trap of implicit patterns would enable the provision of additional insights for graph embedding. But the previous graph-based methods did not encode this sentiment information into the node embedding.    

\begin{figure*}[!t]
\centering
\includegraphics[width=6in]{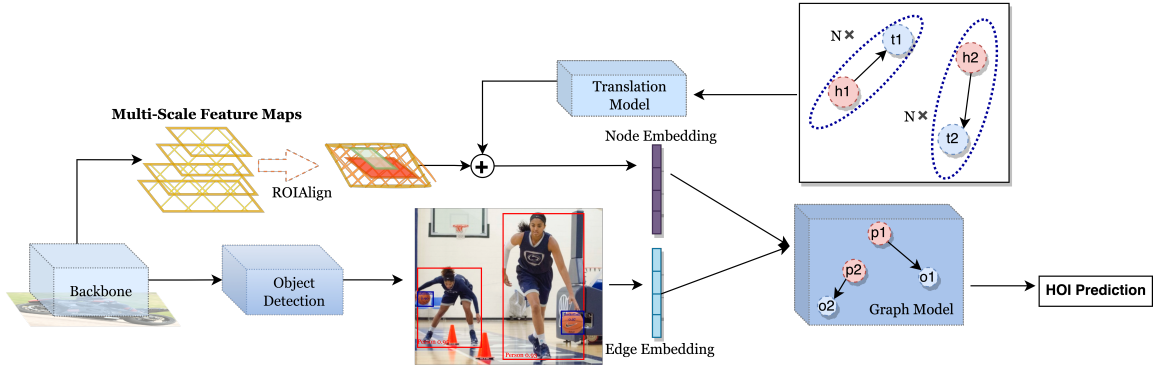}
\caption{This is the framework of the proposed method. Given an image, the backbone module extracts its features, which are then passed into object detection. With extracted feature maps, object detection makes predictions for the instance class, score, and box coordinates. Upon detecting each human and object instance pair, a total of $N \times (h,r,t)$ triplets are generated, with the majority of them being negative. These triplets are subsequently fed into the translational model. The node embedding of the HOI graph model is obtained by concatenating the appearance features and transferring embedding from the translational model, the detection information acquired through object detection is utilized as edge embedding, and the HOI prediction is ultimately produced by the graph model.}
\label{fig_1}
\end{figure*}

In light of the aforementioned limitations, this paper introduces a novel graph-based model for HOI detection named TMHOI: Translational Model for Human-Object Interaction Detection, as illustrated in \hyperref[fig_1]{Figure.2}. Our proposed approach employs the translational model to encode relation semantic features and then integrate them into the node embedding, thus strengthening the link between the appearance feature and the relation semantic feature. Furthermore, unlike previous graph-based methods that constructed node and edge embedding independently, our approach fortifies the connection between nodes and edges by incorporating relation features into node embedding. For instance, consider the example presented in \hyperref[fig_sim]{Figure.1}. Previous graph-based techniques only relied on independent instance features to represent nodes. However, our proposed approach encodes relation features (i.e., the relationship between human-object pairs) into node representations. As a result, the node embedding is able to incorporate relation features and interact more effectively with edge embedding, which also aims to represent relation features.  
This technique integrates the appearance feature and the novel semantics into the node encoding process, imparting supplementary insights to the graph model. In general, to obtain a more meaningful graph representation, TMHOI re-encoded both the edge and node embedding, while the node encoder applied the translational model to do semantic feature extraction. To summarize, our contributions are three-fold:
\begin{itemize}
    \item Our proposed approach involves integrating new semantic representations from the translational model into the node embedding process of the graph-based HOI model, which represents a novel node embedding integration. This is a significant contribution, as it allows the graph-based HOI model to leverage the benefits of the new node and edge embedding representations.
    \item In our proposed method, we have redesigned the node representation to enhance the coherence between the node and edge of the graph neural network. This is achieved by incorporating a relationship feature into the node embedding, which was previously absent in the graph network. By doing so, the node and edge can share interactive features, resulting in an improved prediction ability of the graph neural network.
    \item We modify the knowledge graph embedding model as translation to fit into the graph-based HOI detection framework and enable domain-transfer learning. By leveraging domain-transfer learning, our model has the potential to improve the accuracy of HOI detection in real-world applications, especially in scenarios where labeled HOI detection data is limited.
\end{itemize}
Then reminded of the paper is structured as follows: Section \rom{2} provides a literature review of human-object interaction (HOI) detection. Section \rom{3} outlines our proposed methodology for addressing the formulated problem. In Section \rom{4}, we present a comprehensive comparison and analysis of our approach with state-of-the-art HOI algorithms. The conclusions of our study and the contributions of our proposed methods are presented in Section \rom{5}.

\section{Related Work}
\subsection{Human-Object Interaction Detection}
 Human-Object Interaction Detection could be considered a downstream task of object detection. The advancement of HOI detection is intricately linked with the progress of object detection \cite{girshick2015fast}. More recent methods have adopted deep learning architectures, and many HOI detection models now rely on neural network architectures. Existing work widely-used multi-steam framework proposed by Chao \etal~\cite{chao2018learning}, consists of three streams: a human stream, an object stream, and an interaction stream. Several recent HOI detection algorithms have been proposed that incorporate various improvements.   
 
 HOI detection approaches can be broadly categorized into one-stage and two-stage methods based on their model architecture. Until the recent proposal of QPIC\cite{tamura2021qpic}, two-stage methods have been the predominant approach in HOI detection, with human-pose approaches\cite{wan2019pose}, graph-based approaches\cite{li2019transferable,zhang2021spatially}, and other multi-stream approaches\cite{hou2020visual} leveraging different frameworks and methods for predicting verbs in the head. In addition to the diverse head designs utilized in two-stage HOI detection, the effectiveness of the model is also influenced by the object detector employed. In contrast to two-stage HOI detection, where instances are first detected using an object detector, one-stage approaches directly generate predictions by leveraging the backbone of the object detector without performing explicit instance detection. The majority of one-stage HOI detection methods rely on transformer-based architectures, while others employ interaction point-based methods. One such approach, proposed by Liao \etal~\cite{liao2020ppdm}, entails a pipeline that does not depend on object bounding box but instead directly detects interactions as key points. In recent years, transformer-based HOI detection has gained significant attention and demonstrated promising results, largely due to QPIC's innovative use of transformer-based techniques in HOI detection. Subsequently, several works have followed in the footsteps of QPIC, such as \cite{chen2021qahoi,liao2022gen,wu2022mining}. However, it is not limited to one-stage approaches that have adopted transformer-based architectures, as demonstrated by the use of transformer-based techniques in the two-stage HOI detection method UPT\cite{zhang2022efficient}. Despite the effectiveness of transformer-based methods in achieving good results in HOI detection, their application requires a substantial amount of computational resources, and the resulting predictions are often opaque, posing challenges to subsequent inference work. Furthermore, the relative homogeneity of HOI detection methods in the inference field exacerbates these challenges. Graph-based methods have been proposed as an alternative to address the aforementioned challenges faced by previous methods. Despite the fact that accuracy may be compromised to some extent, graph-based methods strike a good balance between computational efficiency and accuracy. Particularly, graph-based methods show great potential in reducing computation costs as they can effectively encode a graph representation with the aid of knowledge graph embedding models.

\textbf{Graph-based HOI approach} The graph-based HOI approach, which was proposed by Qi \etal~\cite{qi2018learning}, parses the graph neural network\cite{scarselli2008graph} and generates features for HOI detection. The box appearance features serve as the initial values for the node features, which are subsequently updated through an iterative message-passing algorithm. In the following work, Wang \etal~\cite{wang2020contextual} contended that the graph should account for the presence of heterogeneous nodes, human nodes, and object nodes. Gao \etal~\cite{gao2020drg} argue that employing homogeneous nodes within separate sub-graphs: human-centric and object-centric graphs, is beneficial because different nodes are responsible for passing and updating distinct information. Then VSGNet\cite{ulutan2020vsgnet} enhances the features by considering the spatial relationship between the interacting human-object pair and employs graph convolutions to exploit the inherent structural connections within the pair. A recent graph-based HOI detection model, named SCG \cite{zhang2021spatially}, innovatively conditions the messages between pairs of nodes on their spatial relationships. To sum up, the graph model builds it up the CNN-based feature extraction, while CNNs are limited to extracting local features, Graph models can leverage spatial information and capture global dependencies, which is a valuable complement to CNN-based backbones.
\subsection{Knowledge Graph Embedding Model}
Knowledge graph embedding (KGE) is a method that uses low-dimensional vectors to represent the entities and relations in a knowledge graph. This approach makes a good trade-off between model capacity and efficiency and can be applied to a range of downstream tasks, such as link prediction. By using KGE, it becomes possible to effectively capture the underlying structure of the knowledge graph and use it for various purposes. One of the earliest and most popular knowledge graph embedding methods is TransE\cite{bordes2013translating}, which models each relation as a translation operation between the embedding of the head and tail entities. Since then, numerous other embedding methods have been proposed, such as TransH\cite{wang2014knowledge},
which introduce more complex transformations and modeling techniques to capture different relations of the knowledge graph structure. TransH considers the relation vector as a hyperplane and projects the entity vectors onto this hyperplane to compute the score. This approach enables TransH to model one-to-many and many-to-one relationships more effectively than TransE. We adopted TransH as the translational model for this study since it could handle one-to-many human-object interactions. This study is also the pioneer in using knowledge graph embedding (KGE) methods for HOI detection applications since no such methods have been previously employed in this domain.

\section{Methods}
\begin{figure*}[!t]
\centering
\includegraphics[width=4in,height=2in]{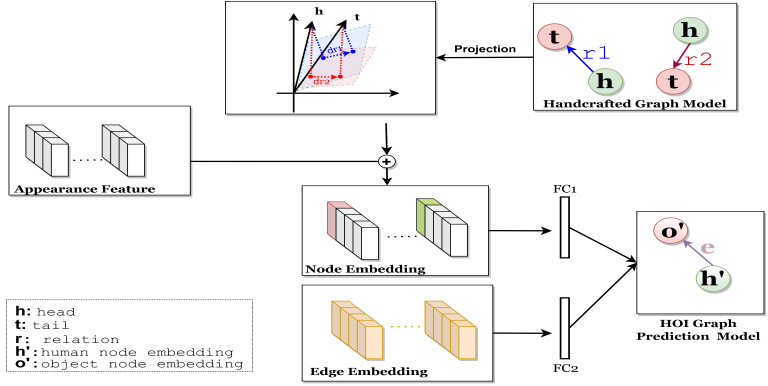}
\caption{The design of the novel semantic graph representations. For every human-object pair, we form N triplets $(h,r_{1},t),\cdots,(h,r_{N},t)$. The objective is to project all relationships between the human and object onto the relation-specific hyperplane $\boldsymbol{w}_{r}$, with the relation-specific translation vector $\boldsymbol{d}_{r}$ indicating the distance between the head and tail embedding in $\boldsymbol{w}_{r}$. This approach aims to reduce the distance between golden triplets and increase the distance between negative triplets. The node embedding integers are the translation feature and appearance feature, and the edge embedding is the spatial feature, pass these graph representations into the FC layers and the HOI graph prediction model sequentially.}
\label{fig_2}
\end{figure*}
In order to enhance the utilization of the graph-based approach, we will employ AdaMixer\cite{gao2022adamixer} as our object detector and develop a more efficient graph head for detecting human-object interactions. The network's general introduction is presented in Section 3.1. While Section 3.2 delves into the design of graph representations. Section 3.3 presents our model training and inference.

\subsection{Algorithm overview}
\label{3.1}
In \hyperref[fig_1]{Figure. 1}, we present a high-level overview of our HOI detection network. This model is comprised of two stages. During the first stage, an object detector is utilized to predict all instances within the image. The predicted set should include $(bbox_{i}, score_{i}, label_{i})$, where $bbox_{i} \in \mathbb{R}^{4}$ represents the $i$-th instance bounding box coordinate which includes the upper left and lower right points $\left[ x_{1},y_{1},x_{2},y_{2} \right]$. The $i$-th confidence score, $score_{i}$, ranges between 0 and 1, while $label_{i} \in M$ represents the predicted label for the $i$-th object. The value of $M$ depends on overall dataset object classes and usually includes the `person' class as part of the object class. 
The low-score instances will be filtered out by Non Max Suppression (NMS)\cite{hosang2017learning} to improve the quality of the detected instances. The threshold value is set to 0.5, and low-quality detected objects are not considered for further analysis in the detection process. 
During the second stage, we extract multi-scale feature maps from the object backbone and use RoIAlign \cite{he2017mask} to obtain the appearance features of a ($bbox_{i}$). Then pair all human and object instances, and classify them as $\mathcal{H}$ set or $\mathcal{O}$ set with the detected label, where $\mathcal{H} = \left\{ label_{i} = ``person" \right\}$ and $\mathcal{O} = \left\{label_{i} \neq ``person" \right\}$. For the pair of the human and object instances, we denote the $label_{h}$ as the human label and $label_{o}$ as the object label. Earlier studies\cite{zhang2021spatially} employed the appearance feature as the node embedding and the spatial pair information as the edge embedding, which were subsequently fed into a graph model. Building on this research, we also feed the node and edge embedding into an HOI graph prediction model. However, we use a novel semantic graph representation, and the HOI graph prediction model is capable of generating HOI predictions based on a novel representation. The graph model we employed was a bipartite graph proposed by Zhang \etal~\cite{zhang2021spatially}.

\subsection{Graph Representation}
\label{3.2}
Our approach involves adopting an \textit{invisible semantic representation} of the human-object pair. This semantic representation could give the graph model some clues to follow the common pattern among different human-object pairs and improve the accuracy of our HOI predictions. \hyperref[fig_1]{Figure 2} illustrates the design of the translational model, which integrates the transfer relation representation onto the node embedding. We embraced the knowledge graph embedding model as the translational model in this research and denoted it as $\mathcal{G_{T}} = (\mathcal{E}, \mathcal{R})$, where $\mathcal{E} \in \mathbb{R}^M$ is the set of entities and $\mathcal{R} \in \mathbb{R}^N$ is the set of the relations. Typically, in the knowledge graph embedding model, $h \in \mathcal{E}$ denotes the head entity, $r \in \mathcal{R}$ denotes a relation type, and $t \in \mathcal{E}$ denotes the tail entity. Each triplet is formed as $(h,r,t) \in \mathbb{R} ^{ \mathcal{E} \times \mathcal{R} \times \mathcal{E} }$. In the latent embedding space, each entity $e \in \mathcal{E}$ and $r \in \mathcal{R}$ are represented as $d$-dimensional vectors. The corresponding embedding representation of the head, relation, and the tail is denoted as \textbf{h, r, t}. The dissimilarity of the triplet $(h, r, t)$ within the embedding space is calculated by the score denoted as $\boldsymbol{s}_{r}(\boldsymbol{h},\boldsymbol{t})$.  The set of triplets is correct in terms of ground truth and is represented by golden triplets $\bigtriangleup$.

Followed by the TransH\cite{wang2014knowledge}, we developed a knowledge graph embedding model that transfers the relationship between humans and objects. The design of the triplet is handcrafted, we would like to map all relations between the human and the object onto the relation-specific space. 
For each relation $r_{i} \in \left\{1,..., N \right\}$, where $N$ denotes the total number of possible relationships. We adopted the $label_{h}$ as the head since all triplets have the same head, the tail will be the $label_{o}$, and the relation is the $r_{i}$. Given a pair between the human and object, we will have $N$ triplets of this pair in the translational model. This translational model enables the diverse distributed representation of the different relations. As illustrated in \hyperref[fig_2]{Figure.3}, for the same head and tail, we project the relation-specific vector $\boldsymbol{d}_{r}$ into the relation-specific hyperplane, instead of the same embedding space. Then denoted \textbf{h}, \textbf{r}, and \textbf{t} as the corresponding embedding representation, the projection is denoted as $\boldsymbol{h}_{\perp}$ and $\boldsymbol{t}_{\perp}$, respectively. If $(h,r,t)$ is a golden triplet, we anticipate that the projection $\boldsymbol{h}_{\perp}$ and $\boldsymbol{t}_{\perp}$ can be linked through a translation vector, $\boldsymbol{d}_{r}$, with a short distance in the relation-specific hyperplane. The score function of the triplet is defined as
\begin{equation}
\boldsymbol{s}_{r}(\boldsymbol{h},\boldsymbol{t}) = \|(\boldsymbol{h} - \boldsymbol{w}_{r}^{T}\boldsymbol{hw}_{r}) + \boldsymbol{d}_{r} - (\boldsymbol{t} - \boldsymbol{w}_{r}^{T}\boldsymbol{tw}_{r})\|_{2}^{2}
\end{equation}

If the triplet is a valid one, we expect that the score $\boldsymbol{s}_{r}(\boldsymbol{h},\boldsymbol{t})$ would be high, whereas it would be low for an invalid one. The translational model encodes the semantic representation of the detected label. During the training of the translational model, the \textbf{h} and \textbf{t} aggregate the knowledge of the head and tail, then we will integer translational knowledge into the graph embedding. Our HOI graph-based approach involves two key types of representations: 1) node embeddings and 2) edge embeddings.

  \textbf{Node Embedding}   
We obtain the instance appearance feature $\boldsymbol{f}$ from the ROIAlign, then passed the ROI pooled feature of the human/object into a MultiLayer Perceptron (MLP) to get a lower dimension. We denoted $\boldsymbol{f}_{h}$ as the human appearance feature and $\boldsymbol{f}_{o}$ as the object appearance feature. Then we formulate the node embedding by concatenating the appearance feature $\boldsymbol{f}$ and entity projection \textbf{h}(or \textbf{t})that we obtain from the translational model. The human node $\mathcal{N}_{h}$ and object node $\mathcal{N}_{o}$ could be written as 
\begin{align}\label{General}
\mathcal{N}_{h} = \sigma(FC_{1}(\boldsymbol{f}_{h} \oplus \boldsymbol{h}))\\
\mathcal{N}_{o} = \sigma(FC_{2}(\boldsymbol{f}_{o} \oplus \boldsymbol{t}))
\end{align}
where $\boldsymbol{f}_{h}$,$\boldsymbol{f}_{o} \in \mathbb{R}^{1024}$ are linear projection of the appearance feature.The size of embedding \textbf{h} and \textbf{t} is determined by the hyper-parameter $k$. FC denotes the fully connected layer, two FC layers have independent weights, allowing for feature projection. $\sigma$ is the ReLU activation function. The node embedding $\mathcal{N}_{h}$,$\mathcal{N}_{o} \in \mathbb{R}^{1024+k}$.  

\textbf{Edge Embedding} We expect the edge embedding could learn the connection between the pair of human and object nodes. Following the approach proposed in SCG\cite{zhang2021spatially}, we encode basic spatial information and pairwise information of bounding boxes and normalize them with respect to the corresponding image dimensions. Additionally, we enhance the pairwise connections by encoding the positions ratio, height ratio, and weight ratio of the pairwise bounding boxes, and including the area ratio of the pairwise bounding box as well. 
  
\subsection{Training and Inference}
\label{3.3}  
In the translational model training, we utilize a margin-based ranking loss to encourage distinction between correct triplets (golden triplets) and incorrect triplets (which are absent in the dataset). A correct triplet receives a positive translaional score, while an incorrect triplet is assigned a negative translation score $\boldsymbol{s}_{r}(\boldsymbol{h},\boldsymbol{t})$. The following rules are used to determine whether a score is positive $\boldsymbol{s}_{r}(\boldsymbol{h},\boldsymbol{t})$ or negative $ \boldsymbol{s}_{r}(\boldsymbol{h'},\boldsymbol{t'})$:
\begin{equation}
\begin{cases}
    \boldsymbol{s}_{r}(\boldsymbol{h},\boldsymbol{t}),\quad & \text{if} \quad (h,t,r) \in \bigtriangleup \\
    \boldsymbol{s}_{r}(\boldsymbol{h'},\boldsymbol{t'}), \quad & \text{if}\quad  (h,t,r)  \notin \bigtriangleup \\
\end{cases} 
\end{equation}
\begin{center}
\begin{table*}[h]
\renewcommand{\arraystretch}{1.3}
\caption{Comparison with state-of-the-art on HICO-DET test set under the default setting. See Section IV for the default setting. The letters `A', `S', `P', `L', and `T' correspond to appearance, spatial, human pose, language, and translational features, respectively. The translation feature refers to the feature obtained from the translational model. Fine-tuning detection involves using a detector that is first trained on the MS-COCO dataset, and then fine-tuned on the HICO-DET dataset. }
\resizebox{\textwidth}{!}{
\setlength{\tabcolsep}{5mm}{
\begin{tabular}{c|c|c|c|c|c c c}
\hline
\multirow{2}{*}{Architecture} & \multirow{2}{*}{Method} & \multirow{2}{*}{Backbone} & \multirow{2}{*}{\begin{tabular}[c]{@{}c@{}}Fine-tuned \\ Detection\end{tabular}} & \multirow{2}{*}{Feature} & \multicolumn{3}{c}{Default} \\ \cline{6-8} 
 &  &  &  &  & \multicolumn{1}{c}{\textit{full(mAP\%)}} & \multicolumn{1}{c}{\textit{Rare(mAP\%)}} & \textit{Non-Rare(mAP\%)} \\ \hline\hline
\multirow{4}{*}{Graph-Based} & RPNN\cite{zhou2019relation} & ResNet-50 & \XSolidBrush  & A+P & \multicolumn{1}{c}{17.35} & \multicolumn{1}{c}{12.78} & 18.71 \\ \cline{2-8} 
 & VSGNet\cite{ulutan2020vsgnet} & ResNet-152 & \XSolidBrush  & A+S & \multicolumn{1}{c}{19.80} & \multicolumn{1}{c}{16.05} & 20.91 \\ 
  \cline{2-8} 
 & SCG\cite{zhang2021spatially} & ResNet-50-FPN & \XSolidBrush  & A+S & \multicolumn{1}{c}{21.69} & \multicolumn{1}{c}{17.69} & 22.88 \\
 \cline{2-8} 
 \rowcolor{lightgray!40}
 & \textbf{TMHOI} & \textbf{ResNet-50-FPN} & \XSolidBrush  & \textbf{A+S+T} & \multicolumn{1}{c}{\textbf{22.61}} & \multicolumn{1}{c}{\textbf{15.87}} & \textbf{24.62} \\ 
 \hline\hline
\multirow{2}{*}{Interaction Points} & PPDM\cite{liao2020ppdm} & Hourglass-104 & \CheckmarkBold & A & \multicolumn{1}{c}{21.73} & \multicolumn{1}{c}{13.78} & 24.10 \\ \cline{2-8} 
 & GGNet\cite{zhong2021glance} & Hourglass-104 & \CheckmarkBold & A & \multicolumn{1}{c}{23.47} & \multicolumn{1}{c}{16.48} & 25.60 \\ \hline
\multirow{3}{*}{Graph-Based} & DRG\cite{gao2020drg} & ResNet-50-FPN & \CheckmarkBold & A+S+L & \multicolumn{1}{c}{24.53} & \multicolumn{1}{c}{19.47} & 26.04 \\ \cline{2-8} 
 & SCG\cite{zhang2021spatially} & ResNet-50-FPN & \CheckmarkBold & A+S & \multicolumn{1}{c}{25.63} & \multicolumn{1}{c}{19.44} & 27.48 \\ \cline{2-8} 
 \rowcolor{lightgray!40}
 & \textbf{TMHOI} & \textbf{ResNet-50-FPN} & \textbf{\CheckmarkBold} & \textbf{A+S+T} & \multicolumn{1}{c}{\textbf{26.95}} & \multicolumn{1}{c}{\textbf{21.28}} & \textbf{28.56} \\ \hline
\end{tabular}}}
\label{table1}
\end{table*}
\end{center}
During the triplet score generation process, the number of negative scores is significantly higher than positive scores. This is because a single human-object pair has only a few interactions at the same time. However, when we developed the translational model, we proposed $N$ triplets to represent $N$ relation-specified relationship. Triplets that not occurring in the dataset are considered negative. However, in order to maintain an equivalent number of positive and negative scores, we randomly select negative triplets in the same quantity as the golden triplets during training. The loss of the translational model is defined as
\begin{equation}
    \mathcal{L}_{T} = \sum_{\substack{(h,r,t) \in \bigtriangleup}}\sum_{\substack{(h',r',t') \in \bigtriangleup'}}\left[\boldsymbol{s}_{r}(\boldsymbol{h},\boldsymbol{t})+\delta-\boldsymbol{s}_{r}(\boldsymbol{h}',\boldsymbol{t}')\right]_{+}
\end{equation}
where $[x]_{+} \triangleq max(0,x)$, we applied the hyper-parameter $\delta$ as the margin to divide the set of golden triplets and negative triplets. 
By applying node and edge embedding, the Spatial-Semantic embedded nodes and edges will pass through the graph network (following the SCG \cite{zhang2021spatially} approach) and get the verb prediction scores set $\mathcal{C} = \left(c_{1},...,c_{j},...c_{\mathcal{H} \times \mathcal{O}} \right)$, where $\mathcal{H} \times \mathcal{O}$ are the total number of all possible pairs of humans and objects indexed by $j$. To supervise the existence of the corresponding human-object pair detection in all possible pairs, we define
\begin{equation}\label{eq:pscore}
    p_{j} = (s_{j}^{h})^{\lambda} \dot (s_{j}^{o})^{\lambda}
\end{equation}

\noindent where $(s_{j}^{h})^{\lambda}$ and $(s_{j}^{o})^{\lambda}$ denote the detection scores for humans and objects in $j$-th pair. The $\lambda$ is set to 1 during training and set to 2.8 during inference to overcome the effect of overconfidence detection scores. From \eqref{eq:pscore}, we obtain the interactiveness scores set $\mathcal{P} = (p_{1},...,p_{j},...p_{\mathcal{H} \times \mathcal{O}})$ to indicates the detection probability of each pair.
With $\mathcal{P}$ and HOI verb prediction scores $\mathcal{C}$ during the HOI graph model training. We have the final action score as follows: 

\begin{equation}
    \mathcal{V} = \mathcal{P}\cdot\mathcal{C}
\end{equation}

Additionally, we have another supervisor $\mathcal{W} = (w_{1},...,w_{j},...w_{\mathcal{H} \times \mathcal{O}})$
$\mathcal{E}$ indicates interaction probability between all graph pairs. It will help to identify whether the $j$-th pair
exists. We use binary focal loss\cite{lin2017focal} which formulated as:
\begin{equation}
    FL(\hat{y},y)
    \begin{cases}

    -\beta(1-\hat{y})^{\gamma}log(\hat{y}), \quad & \text{if} \quad y=1\\
    -(1-\beta)\hat{y}^{\gamma}log(1-\hat{y}), \quad & \text{if} \quad y=0

    \end{cases}
\end{equation}
\noindent where $\hat{y}$ is the prediction value between 0 and 1, and ground truth $y \in \left\{0,1\right\}$. So we define the interactive loss as $\mathcal{L}_{W} = FL(\hat{\mathcal{W}}, W)$, where $W$ is the ground truth that if there is a pair exists, will be 1, otherwise 0. We also define verbal classification loss as $\mathcal{L}_{V} = FL(\hat{\mathcal{V}},V)$, where $V$ is the ground truth of verb action. The objective function of our proposed network could be expressed as:
\begin{equation}
    \mathcal{L} = \mathcal{L}_{T} + \mathcal{L}_{W}+\mathcal{L}_{V} + 
\end{equation}

\section{Experiments}

\subsection{Dataset and Metric}
\textbf{Dataset} Our model was tested on the HICO-DET\cite{chao2018learning} datasets. The HICO-DET is a large-scale benchmark that contains 37,633 training and 9,546 test images, each annotated with a set of human-object interactions and the corresponding object bounding boxes. The interaction types are categorized into 600 classes, and the dataset contains a total of 80 object classes and 117 different action classes. The distribution of pairs across interaction classes is heavily imbalanced, with a long-tailed distribution. Specifically, there are 47 interaction categories that have only one training example. 

\textbf{Metric}  we use the mean average precision (mAP) metric to assess the accuracy of the predicted human-object interaction (HOI) pair. A predicted HOI instance is considered a true positive if the intersection over union (IoU) between the predicted human bounding box and the ground-truth human bounding box is greater than 0.5, and the IoU between the predicted object bounding box and the ground-truth object bounding box is also greater than 0.5. For every configuration, we provide the mean average precision (mAP) for three sets of HOI classes: the full set of 600 HOI classes (full), a set of 138 HOI classes with fewer than 10 training instances (rare), and a set of 462 HOI classes with 10 or more training instances (non-rare). The default setting for evaluating HOI (human-object interaction) detection involves testing on the entire set of images, including those that do not contain the object being targeted. 

\textbf{Implement Details} We utilize the AdaMixer as the object detector with the pre-trained ResNet50-FPN\cite{he2016deep} backbone. At the end of the first stage, we filter out the detections if their detected score is lower than 0.2, then perform non-maximum suppression(NMS) in a batched fashion. While training, we exclude pairs of the same person but retain pairs involving different individuals, since we regard the person as an object category as well. While in the second stage, we utilize the multi-scale feature maps obtained from various pyramid levels by employing FPN (Feature Pyramid Network) \cite{lin2017feature}. Subsequently, ROIAlign is applied to extract the box feature with an output size of $7 \times 7$. The box features are projected into a 1024-dimensional space using a two-layer MLP, combined with the transfer semantic feature from the translational model to form a node feature. Regardless of the size of the new node representation, it will map to a 1024-dimensional space before being passed into the HOI graph model. The edge representation undergoes the same operation, wherein the spatial feature is transformed into a 1024-dimensional space through a three-layer MLP. Finally, we set $\beta = 0.5$ and $\gamma = 0.2$ for the focal loss, and $\delta = 4$ for the margin-based rank loss.

We train the graph models for 12 epochs using the AdamW optimizer \cite{loshchilov2017decoupled} with a batch size of 16, incorporating a momentum of 0.9 and weight decay of $10^{-4}$. Also, the object detection was trained with the AdamW optimizer, and the learning rate is set to $2.5 \times 10^{-5}$, with 36 epochs and a weight decay of $10^{-4}$. 

\subsection{Comparison and Analysis}

The comparison of our results with state-of-the-art methods on the HICO-DET dataset was shown in the \hyperref[table1]{Table 1}. Our best model for comparison with other state-of-the-art graph-based approaches is TMHOI with $k=50$. As most outperforming approaches require significant computational resources and longer convergence times due to their reliance on transformer architecture in the prediction head, we limited our comparison to graph-based HOI detection methods. While these models deliver good performance, they come with substantial training costs. Graph-based approaches, however, offer a more efficient and computationally tractable alternative, leveraging their structure to achieve high levels of accuracy without the same resource demands.    

The present study examines the performance of the TMHOI algorithm in comparison to other graph-based approaches, both with and without the utilization of a fine-tuned detector. Our findings demonstrate that TMHOI outperforms other graph-based methods by achieving a mean average precision (mAP) of 0.92, without utilizing the fine-tuned detector. With the utilization of the fine-tuned detector, TMHOI exhibits superior performance, achieving a significant 1.32 mAP increase over the state-of-the-art method. Additionally, when compared to the interaction point approach, TMHOI outperforms with a margin of 3.48 mAP, further attesting to its effectiveness in this domain. These results emphasize the efficacy of the TMHOI algorithm and highlight its potential in HOI detection tasks.  

The performance of HOI approaches varies across different settings, revealing notable variations based on the use of a fine-tuned detector versus a detector without fine-tuning and the different feature utilization methods. This underscores the critical role of the object detector in two-stage HOI detections, as all works are built on the detections. In addition, the feature used for the HOI model is also vital, with more information utilized in the graph leading to better performance from the model. It is noted that the approach used in TMHOI, which utilized not only the appearance and spatial features but also the translation feature, resulted in a higher mAP score of 1.32 compared to SCG\cite{zhang2021spatially} that only applied appearance and spatial features. Nonetheless, incorporating additional features into the model leads to an upsurge in computational expenses. To mitigate this issue, we propose the application of Knowledge Graph Embedding (KGE) for feature translation. Notably, this method incurs minimal costs while providing a highly advantageous feature. In contrast to DRG\cite{gao2020drg}, which leverages word embeddings of detected labels as a 300-dimensional semantic feature, TMHOI requires only a 50-dimensional semantic feature that is translated using Knowledge Graph Embedding (KGE). In conclusion, our proposed approach not only provides a novel feature but also achieves a higher mAP score in a simple and efficient manner(KGE).
\begin{table}[!t]
\renewcommand{\arraystretch}{1.3}
\caption{Ablation study for the different embedding sizes $k$ of the translational model, the `-' means we didn't utilize the translational model, kindly refer to the Section 4.2}
\label{table_2}
\centering
\begin{tabular}{c|c|c|c}
\hline
\bfseries $k$ & \bfseries full(mAP\%) & \bfseries rare(mAP\%) & \bfseries non-rare(mAP\%)\\
\hline\hline
- & 26.08 & 19.93 & 27.92\\
30 & 26.45 & 20.63 & 28.19\\
\textbf{50} & \textbf{26.95}& \textbf{21.28}& \textbf{28.56}\\
70 & 26.48& 20.20& 28.26\\
100 & 26.56 & 21.03 & 28.21\\
\hline
\end{tabular}
\end{table}
\subsection{Ablation Study}
The importance of the embedding size cannot be overstated in knowledge graph embedding models. The size determines the number of dimensions in the continuous vector space utilized to represent entities and relations as low-dimensional embeddings. In the translational model, each relation is linked to a hyperplane in the embedding space, and entity embeddings are projected onto these hyperplanes to generate their representations. To ensure exceptional performance on downstream tasks like link prediction and entity classification, selecting the ideal embedding size is critical. If the embedding size is too small, the model's expressiveness may be limited, and performance may suffer. Conversely, a too-large embedding size may result in overfitting and slower training. Therefore, it is imperative to conduct an ablation study to examine the impact of different embedding sizes on the model's performance.

We investigate the impact of different embedding sizes on the performance of a knowledge graph model and show the result in the \hyperref[table_2]{Table 2}. Specifically, we train a knowledge graph embedding model as a translational model on a large-scale image dataset and vary the embedding size from 30 to 100. The finding is that an embedding size of 50 yields the best performance. It is possible that an embedding size of 50 strikes a balance between the capacity of the translational model to capture the complexity of relation knowledge and the potential for overfitting. Notably, without utilizing the translational model to incorporate the supplementary feature, the model performs poorly, exhibiting an mAP point difference of nearly one compared to $k=50$.

The proportion between the size of the relation embedding and the appearance feature in the node embedding is an important consideration in node embedding. As the size of the translation embedding increases, the proportion of the relation feature in the node embedding also increases, as the node embedding is a composite of the translation embedding and appearance feature of the node embedding. However, if the proportion of the relation feature becomes too large, it may adversely affect the final performance of the model. Thus, finding the appropriate balance between the appearance and relation features in the node embedding is a crucial task for graph-based models. In our experiments, an embedding size of $k=50$ resulted in the best performance, suggesting that this particular size strikes a favorable balance between the appearance and relation features in the node embedding.

\section{Conclusion}
This paper presents a novel graph-based approach for human-object interaction detection, which employs a knowledge graph embedding model as the translational model to embed the relation feature and integrate it into the node embedding. By doing so, our approach enhances not only the node representation but also the consistency between the node embedding and edge embedding, leading to improved detection performance. This approach provides a new avenue for researchers to explore the use of knowledge graph embedding models in graph-based approaches for human-object interaction detection.   

Future work could explore improved methods for representing knowledge graph embeddings in graph-based approaches. Such methods have the potential to significantly reduce the computational requirements of graph models, making them more efficient and practical for real-world applications. One area in which this could be particularly useful is in transfer learning \cite{lee2017transfer,9336290}for human-object interaction detection. Many current datasets used for HOI detection are not representative of the complex interactions that occur in real-world settings. As such, it can be difficult to apply models trained on these datasets to real-world scenarios. To address this, we will explore more effective translational models for graph representation, with the aim of improving transfer learning for HOI detection. This research could ultimately lead to more practical and accurate HOI detection models that can be applied in a range of real-world contexts.

\section*{Acknowledgment}
This work was supported by the College of Arts and Sciences and the Department of Computer Science at Bowling Green State University. The authors express their sincere gratitude to the colleges who contributed to this work. Specially, the authors thank Dr. Qing Tian, the Assistant Professor of Computer Science at Bowling Green State University, who generously provided computational resource and valuable advice.
\bibliographystyle{plain}
\bibliography{refs}

\newpage

\section{Biography Section}
\vspace{-20pt}



\begin{IEEEbiography}[{\includegraphics[width=1in,height=1.25in,clip,keepaspectratio]{lijing_zhu.pdf}}]{Lijing Zhu}
is a Data Science Ph.D. student at Bowling Green University, having previously obtained her Master's degree in Analysis from the same institution. Her research interests include Human-Object Interaction Detection and Knowledge Graph Embedding. She is dedicated to pursuing excellence in her field and contributing to its advancement. Her peers and mentors appreciate her technical skills and collaborative approach, and she is excited to continue exploring the possibilities of Data Science in the future.
\end{IEEEbiography}
\vspace{-40pt}
\begin{IEEEbiography}[{\includegraphics[width=1in,height=1.25in,clip,keepaspectratio]{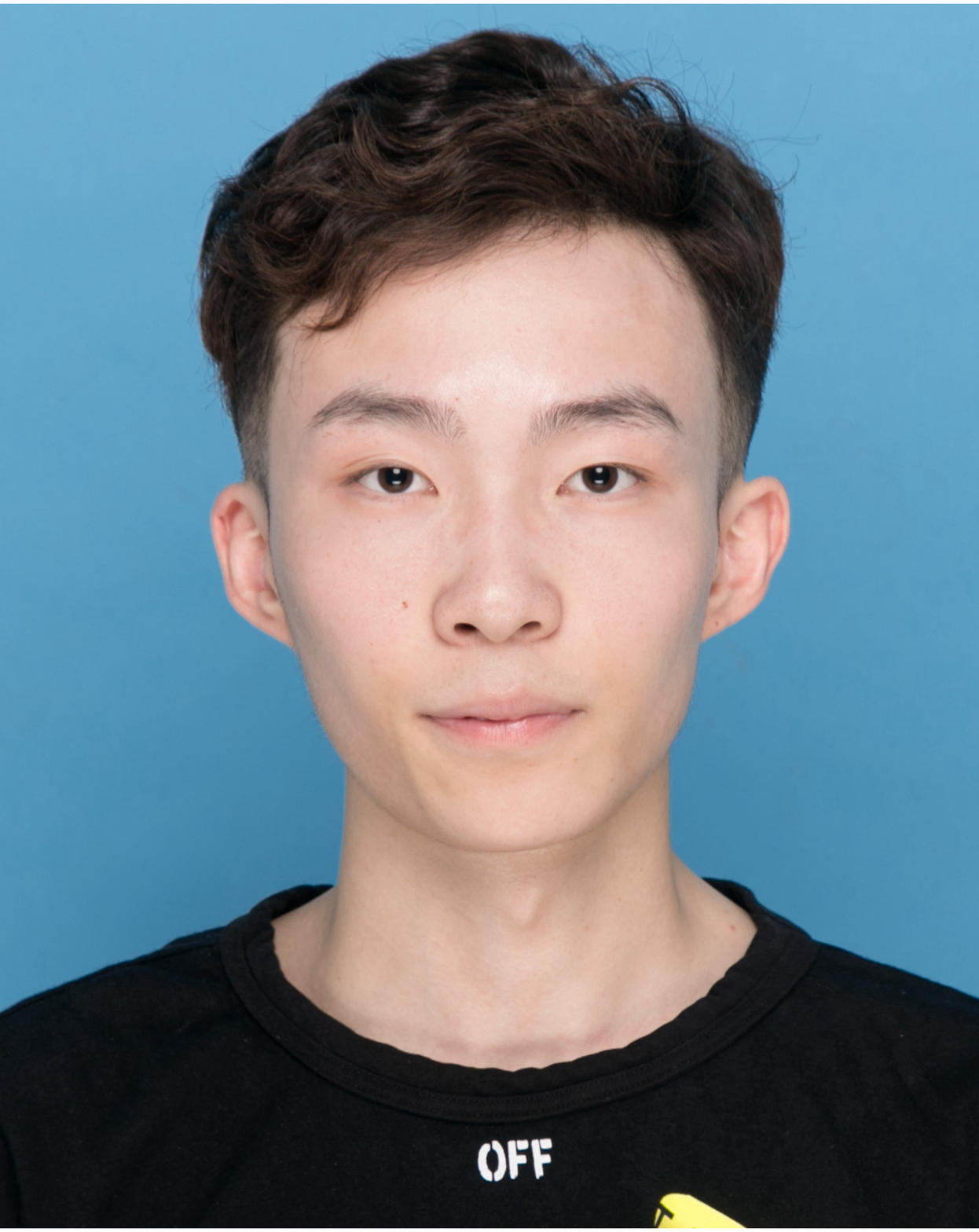}}]{Qizhen Lan}
is currently pursuing the Ph.D. degree in Data Science at Bowling Green State University, Bowling Green, OH, USA. He received the Master’s degree in Intelligence and Analytics in 2019 and the Bachelor’s degree in Business Analytics in 2018, both from Bowling Green State University, Bowling Green, OH, USA. He received the Bachelor's degree in Information Management and Information Systems in 2018 from Tiangong University, Tianjin, China. His research interests lie in object detection, knowledge distillation, and deep network pruning.
\end{IEEEbiography}
\vspace{-20pt}
\begin{IEEEbiography}
[{\includegraphics[width=1in,height=1.25in,clip,keepaspectratio]{DrAlvaro.pdf}}]{Alvaro Velasquez} is a visiting professor of computer science at the University of Colorado Boulder and a program manager in the Innovation Information Office (I2O) of the Defense Advanced Research Projects Agency (DARPA), where he currently leads programs on neuro-symbolic AI. Before that, Alvaro oversaw the machine intelligence portfolio of investments for the Information Directorate of the Air Force Research Laboratory (AFRL). Alvaro received his PhD in Computer Science from the University of Central Florida in 2018 and is a recipient of the National Science Foundation Graduate Research Fellowship Program (NSF GRFP) award, the University of Central Florida 30 Under 30 award, a distinguished paper award from AAAI, and best paper and patent awards from AFRL. He has co-authored over 60 papers and two patents and serves as Associate Editor of the IEEE Transactions on Artificial Intelligence. His research has been funded by the Air Force Office of Scientific Research.
\end{IEEEbiography}
\vspace{-20pt}
\begin{IEEEbiography}
[{\includegraphics[width=1in,height=1.25in,clip,keepaspectratio]{DrSong.pdf}}]{Houbing Song} (M’12–SM’14-F’23) received the Ph.D. degree in electrical engineering from the University of Virginia, Charlottesville, VA, in August 2012.

He is currently a Tenured Associate Professor, the Director of NSF Center for Aviation Big Data Analytics (Planning), and the Director of the Security and Optimization for Networked Globe Laboratory (SONG Lab, www.SONGLab.us), University of Maryland, Baltimore County (UMBC), Baltimore, MD. Prior to joining UMBC, he was a Tenured Associate Professor of Electrical Engineering and Computer Science at Embry-Riddle Aeronautical University, Daytona Beach, FL. He serves as an Associate Editor for IEEE Transactions on Artificial Intelligence (TAI) (2023-present), IEEE Internet of Things Journal (2020-present), IEEE Transactions on Intelligent Transportation Systems (2021-present), and IEEE Journal on Miniaturization for Air and Space Systems (J-MASS) (2020-present). He was an Associate Technical Editor for IEEE Communications Magazine (2017-2020). He is the editor of eight books, the author of more than 100 articles and the inventor of 2 patents. His research interests include cyber-physical systems/internet of things, cybersecurity and privacy, and AI/machine learning/big data analytics. His research has been sponsored by federal agencies (including National Science Foundation, US Department of Transportation, and Federal Aviation Administration, among others) and industry. His research has been featured by popular news media outlets, including IEEE GlobalSpec's Engineering360, Association for Uncrewed Vehicle Systems International (AUVSI), Security Magazine, CXOTech Magazine, Fox News, U.S. News \& World Report, The Washington Times, and New Atlas. 

Dr. Song is an IEEE Fellow, an ACM Distinguished Member, and an ACM Distinguished Speaker. Dr. Song is a Highly Cited Researcher identified by Clarivate™ (2021, 2022) and a Top 1000 Computer Scientist identified by Research.com. He received Research.com Rising Star of Science Award in 2022 (World Ranking: 82; US Ranking: 16). Dr. Song was a recipient of 10+ Best Paper Awards from major international conferences, including IEEE CPSCom-2019, IEEE ICII 2019, IEEE/AIAA ICNS 2019, IEEE CBDCom 2020, WASA 2020, AIAA/ IEEE DASC 2021, IEEE GLOBECOM 2021 and IEEE INFOCOM 2022.
\end{IEEEbiography}
\vspace{-20pt}
\begin{IEEEbiography}[{\includegraphics[width=1in,height=1.25in,clip,keepaspectratio]{Kamal.pdf}}]{Kamal Acharya}{\space}(Graduate Student Member, IEEE) received his Engineering degree in Electronics and Communication Engineering from Tribhuvan University, Kathmandu, Nepal in 2011 and Masters degree in Information System Engineering from Purbanchal University, Kathmandu, Nepal in 2019. Currently, he is pursuing PhD. in Electrical Engineering and Computer Scince from Embry-Riddle Aeronautical University, Daytona Beach, FL.

He has been involved in teaching profession for about 7 years in the various universities of Nepal, Tribhuvan University and Purbanchal Univesity were among few of them. He is mainly associated with the courses like programming(C,C++,Python), Computer Networks and Computer Architecture. He is working as Graduate Research Assistant in Embry-Riddle Aeronautical University. He is also serving as an reviewer for IEEE Transactions on Artificial Intelligence (TAI) and IEEE Transactions on Intelligent Transportation Systems. His preferred areas of research are Natural Language Processing(NLP), Deep Learning and Reinforcemnt Learning.
\end{IEEEbiography}
\vspace{-20pt}
\begin{IEEEbiography}[{\includegraphics[width=1in,height=1.25in,clip,keepaspectratio]{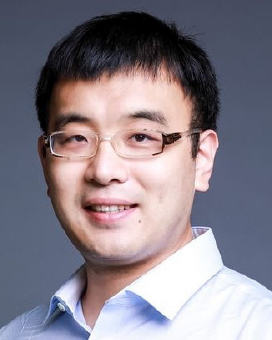}}]{Qing Tian}
received the BEng degree in computer science and engineering from Yanshan University, Qinhuangdao, Hebei, China in 2011. He received the MEng and PhD degrees in electrical engineering in 2013 and 2021, respectively, from McGill University, Montreal, QC, Canada. He is currently an assistant professor at Bowling Green State University, Bowling Green, OH, USA. From 2019 to 2020, he was an applied scientist intern with Amazon.com, Inc. (Visual Search \& AR) at Palo Alto, CA, USA. From 2013 to 2014, he worked as an software developer at Nakisa Inc., Montreal, QC, Canada. His research interests include autonomous driving visual perception, deep network compression, efficient neural architecture search, and adversarial AI. His current research in efficient and robust self-driving perception is supported by the National Science Foundation.
\end{IEEEbiography}
\vspace{-20pt}
\begin{IEEEbiography}[{\includegraphics[width=1in,height=1.25in,clip,keepaspectratio]{SNiu.pdf}}]{Shuteng Niu}
(sniu@bgsu.edu) is a is an Assistant Professor of Computer Science at Bowling Green State University, Bowling Green, OH. Before joining BGSU, he was a Postdoc Research Fellow at School of Biomedical Informatics UTHealth, Houston, TX. Before that, He received his Ph.D. degree in Computer Science at Embry-Riddle Aeronautical University, Daytona Beach, FL. His research areas are machine learning, deep transfer learning, graph representation, and biomedical informatics.
\end{IEEEbiography}

\end{document}